\documentclass{article}

\usepackage{amsmath}
\usepackage{authblk}
\usepackage{graphicx}
\usepackage{longtable}
\usepackage[colorlinks=true,linkcolor=blue,citecolor=blue,urlcolor=blue]{hyperref}
\usepackage{graphicx} 
\usepackage{xcolor}
\usepackage[dvipsnames]{xcolor}
\usepackage{url}
\usepackage{float}
\usepackage{array}
\usepackage[top=0.5in, bottom=0.5in, left=0.75in, right=0.75in]{geometry}
\usepackage{svg}
\usepackage{tikz}
\usepackage{pgfplots}
\usepackage{pgfplotstable}
\usetikzlibrary{shapes.geometric, arrows, positioning}
\usepackage{hyperref}
\usepackage[style=authoryear, maxcitenames=2, uniquename=false, backend=biber]{biblatex}
\pgfplotsset{compat=1.18}

\begin{document}
\title{Decoders Laugh as Loud as Encoders}
\author[1]{Eli Borodach}
\author[1]{Raj Dandekar}
\author[1]{Rajat Dandekar}
\author[1]{Sreedath Panat}

\affil[1]{Vizuara AI Labs}
\date{September 2025}

\maketitle
\begin{abstract}
From the dawn of the computer, Allen Turing dreamed of a robot that could communicate using language as a human being. The recent advances in the field of Large Language Models (LLMs) shocked the scientific community when a single model can apply for various natural language processing (NLP) tasks, while the output results are sometimes even better than most human communication skills. Models such as GPT, Claude, Grok, etc. have left their mark on the scientific community. However, it is unclear how much these models understand what they produce, especially in a nuanced theme such as humor. The question of whether computers understand humor is still open (among the decoders, the latest to be checked was GPT-2). We addressed this issue in this paper; we have showed that a fine-tuned decoder (GPT-4o) performed (Mean F1-macro score of 0.85) as well as the best fine-tuned encoder (RoBERTa with a Mean of F1-score 0.86).
\end{abstract}

\section{Introduction}
\paragraph{\textnormal{There are many types of jokes, and therefore humor can be different. Also, it diverges from language to language since there is a cultural difference between nations. Therefore, we have decided to stick to English humor only. We had divided English Humor into five main categories: absurdity, dark, irony, wordplay (including puns and any other form of words made in a special way to make us laugh), and social commentary (such as in \autocite{kasu2025deceptive}). We added another category of negative examples such as regular sentences that aren't supposed to be funny, and we tagged them as no-joke.}}
\paragraph{\textnormal{We decided to use LLMs to classify each joke to each category, assuming the classification will be good means the LLM understand the jokes. We used three types of LLMs: Encoders, Encoders2Decoders (zero-shot and few-shots learning only), and Decoders (zero-shot, few-shots, and fine-tuned learning). Since the humor dataset categories examples were biased, we decided to weigh scarce categories as abundant categories, therefore we used f1-macro score as the metric for evaluation.}}
\paragraph{\textnormal{
Encoders were trained on a training dataset and evaluated on validation dataset. Since we were lacking data, we used 20 epochs to Fine-Tune the Encoders, and we picked the iteration with the best f1-macro score. Encoders2Decoders were tested only with zero-shot learning and few-shot learning. Decoder (GPT-4o) was fine-tuned using OpenAI API.}}
\paragraph{\textnormal{The results were surprising, Fine-tuned Decoder model is on par with a Fine-tuned Encoder. While it's important to note that decoders mission is to generate text, they weren't "born" to classify. And yet they perform pretty much as the best model found so far: RoBERTa \autocite{faraj2021sarcasmdet}.}}

\section{Literature Review}
Humor serves the human society in order to mitigate the pain from mundane problems, a better way to communicate and convey a message, to make someone like you, or even in the end of the day to provide us with a better feeling (as Charlie Chaplin once said: "A day without laughter is a day wasted."). However, what makes us laugh is studied for a few decades, and the ability to "grasp" the joke is considered to be one of the top enigmas in human's cognition. Considering this, it's clear that this "human" ability is considered hard to "grasp" for a robot. according to a review \autocite{review2024until2022} There are two stages in humorous classification: feature selection and algorithm selection. These two options are going to be covered thoroughly in the next two sub-sections: \autoref{subsec:features_extraction} and  \autoref{subsec:algorithms}

\subsection{Features Extraction}
\label{subsec:features_extraction}
During time there were many approaches for how to get good features for humorous classification. These approaches were divided into three main categories:
\begin{itemize}
\item \textbf{\underline{Feature Engineering}} - Old school methods which were highly explored during the time of classical Machine Learning (ML).
\begin{itemize}
\item \textbf{Ambiguity Detection} - a word that may have more than one meaning can help to identify a joke \autocite{morales2017identifying}, \autocite{reyes2009analysis}, \autocite{hasan2021humor}, \autocite{attardo1994linear}, \autocite{bucaria2004lexical}, \autocite{van2018homonym}, \autocite{diao2020crga, diao2020homographic}, \autocite{yang2015humor}, \autocite{kao2016computational}
\item \textbf{Incongruity Detection} - it's when you have one line of thought interrupted by a different line of thought in the same sentence \autocite{mahajan2017svnit, mahajan2020humor, ziser2020humor, barbieri2014automatic, mihalcea2010computational, liu2018exploiting, liu2018modeling, zhang2017investigations, zhang2018research, morales2017identifying, kamal2019self, yang2015humor, van2018homonym, zhang2014recognizing}
\item \textbf{Emotion-Based Detection} - certain words can imply about the narrator's attitudes, feelings, modes etc. \autocite{reyes2012humor, diao2017homographic, zhang2017investigations, zhang2018research, ortega2018uo}
\item \textbf{Unexpectedness Detection} - a joke can touch a taboo issue or be absolutely about absurd situations - which will make us laugh \autocite{reyes2012humor, barbieri2014automatic, morales2017identifying}
\item \textbf{Subjectivity} - different people find different jokes funny \autocite{zhang2017investigations}. This happens when the joke contains language which implies different beliefs, speculations, criticisms, opinions, and evaluations \autocite{wiebe2006word}. This kind of feature can predict if it contains humor or not \autocite{zhang2017investigations, zhang2018research, liu2018exploiting, liu2018modeling, yang2015humor, kamal2019self, ziser2020humor}
\item \textbf{Negation} - Funny texts usually include negative word such as "not", "no", "isn't", "doesn't", "don't", denials, or any other type of other word with negative meaning 
\autocite{mihalcea2007characterizing, castro2016joke, van2018homonym, zhang2018research, ortega2018uo, shahaf2015inside, yang2019predicting, sjobergh2007recognizing}.
\end{itemize}
\item \textbf{\underline{Automated Features}} - Features that are extracted without manual work, however too many features may end up with curse of dimensionality, therefore researchers use feature selection. 
\begin{itemize}
\item \textbf{Word Embeddings} - words are being represented using dense-vectors which contain the semantic meaning of the word, Words with close meaning have a higher cosine score result 
\autocite{goldberg2017neural}. 
Choosing the right features can improve jokes recognition \autocite{liu2018modeling, hasan2019ur, vadehra2017uwav, shahaf2015inside, diao2017homographic, chen2021through, bertero2016multimodal, indurthi2017fermi, gu2019development, jensen2020buhscitu, kamal2019self, ren2021abml}.
\item \textbf{Bag-of-Words} - This method counts words in the text, without giving any importance for the order of the word or its grammar \autocite{goldberg2017neural}. This end ups with long and sparse vectors. This may be helpful since certain words can be good for classifying the text as humorous or not-humorous \autocite{ermilov2018stierlitz}. Semantics of the text are an important feature which can be seen using this feature extraction and can help to classify the text \autocite{indurthi2017fermi, yang2015humor, ermilov2018stierlitz, yatsu2018comparison, khandelwal2018gender}.
\item \textbf{Part of Speech (POS)} - the frequencies such as nouns, pronouns, verbs, and modifiers for example can predict whether the text is funny or not. It was also interesting to see that the number of appearances or length of the triplet “noun-verb-adjective” also had been a good predictor. 
It was found also useful since for example, there is a correlation between positions of pun words and location of the grammatical tagging 
\item \textbf{Acoustic-prosodic} - When you have voice recording you can detect humorous parts according to the way they are said. However, in this research paper we research only textual information, therefore this type of feature extraction will be ignored. 
\autocite{diao2017homographic, diao2020homographic, diao2020crga, zhang2014recognizing, skalicky2016linguistic, van2018homonym, shahaf2015inside, ren2021abml, ortega2018uo, mahajan2020humor, reyes2009analysis, reyes2012humor, kamal2019self, ermilov2018stierlitz, bucaria2004lexical}.
\end{itemize}
\item \textbf{\underline{Lexical Features}} - a lexical resource consists from language dictionaries, and special addition of data that can contribute to the field of NLP. It contains words, sub-words, and collocations. Each of the fields may have phonological and semantic relations, such as synonyms, spelling, etc. The most commonly used lexical resource is WordNet, although some research findings pointed to the fact that this dictionary is still lacking  \autocite{mihalcea2005making}.
\begin{itemize}
\item \textbf{Alliteration} - Is the case when you have two or more adjacent words, which has the same vocal beginning, which causes the sentence to be unexpected and funny. \autocite{mihalcea2005making, mihalcea2010computational, yang2015humor, morales2017identifying, zhang2017investigations, zhang2018research, kamal2019self, liu2018exploiting, liu2018modeling, sjobergh2007recognizing, ortega2018uo, van2018homonym, bucaria2004lexical, zhang2014recognizing}.
\item \textbf{Antonymy} - When two or more opposite semantic-words are contained in the same sentence, that can make the sentence humorous. This type of feature is being extracted using a dictionary (usually WordNet Antonyms) \autocite{mihalcea2005making, castro2016joke, zhang2018research, vadehra2017uwav, diao2017homographic, van2018homonym, ortega2018uo}. 
\item \textbf{Frequency} - Frequency of certain words, or frequency of certain words that comes together, can suggest whether the sentence is humorous or not, since these words or words combination are common at jokes, while they are rare for regular texts \autocite{skalicky2016linguistic, diao2020crga, diao2020homographic, adams2012identification, barbieri2014automatic, reyes2009humor, gu2019development, kamal2019self, westbury2019wriggly, ermilov2018stierlitz, skalicky2016linguistic, mahajan2017svnit, raz2012automatic}.
\item \textbf{Polarity} - A negative word in a positive sentiment sentence can make us laugh, as well as the other way around \autocite{reyes2012humor, diao2017homographic, liu2018exploiting, liu2018modeling, van2018homonym, reyes2009humor, reyes2012humor, kamal2019self, zhang2014recognizing, yang2015humor, yang2019predicting, mihalcea2007characterizing, skalicky2016linguistic, zhang2017investigations, zhang2018research}. 
\item \textbf{Sentiment-based} - Sentiment is a mental state often shaped by feelings and emotional response, and is commonly used to create humorous elements \autocite{barbieri2014automatic, liu2018modeling, hasan2021humor, ortega2018uo, westbury2019wriggly, shahaf2015inside, van2018homonym, zhang2014recognizing, yang2015humor, yang2019predicting}. 
\item \textbf{Adult slang} - Taboo themes or sexual slang can be a good predictive for humor \autocite{van2018homonym, castro2016joke, mihalcea2005making, yang2015humor, reyes2009humor, ortega2018uo, sjobergh2007recognizing}.
\item \textbf{Written-spoken} - sometimes colloquial words are used are used in a written text. That might be a strong characteristic of a humorous text \autocite{mahajan2017svnit, mahajan2020humor, barbieri2014automatic}. 
\item \textbf{Rhyme} - When two words have the same phonetic ending, but they mean different things, the sentence can be perceived as funny \autocite{zhang2017investigations, zhang2018research, mihalcea2010computational, van2018homonym, yatsu2018comparison, yang2015humor, morales2017identifying, bucaria2004lexical, mihalcea2005making, liu2018exploiting, liu2018modeling, sjobergh2007recognizing, zhang2014recognizing}
\item \textbf{Stylistic} - data of Stylistic characteristics encompass various textual and formatting elements, including word and text length metrics, average word size, connective terms, emojis, hash symbols, web links, referential expressions, abbreviations, question-forming pronouns, capitalized text, bracketed content, highlighted terms, interrogative elements, and punctuation frequency including exclamation points and quotation marks. These features can predict whether the text is humorous or not \autocite{barbieri2014automatic, adams2012identification, castro2016joke, westbury2019wriggly, diao2017homographic, purandare2006humor, reyes2009humor, ermilov2018stierlitz, jensen2020buhscitu, zhang2014recognizing, liu2018exploiting, khandelwal2018gender, raz2012automatic, mahajan2017svnit, mahajan2020humor, ortega2018uo}.
\item \textbf{Human Centeredness} - knowledge such as persons, socials groups, social relations, and personal pronouns can be a good predicative measure for humorous texts \autocite{mihalcea2007characterizing, sjobergh2007recognizing, ortega2018uo, reyes2009humor, shahaf2015inside, yang2019predicting}. 
\item \textbf{Similarity} - Two types of similarity might occur in funny text: syntactic similarity which is focused on the POS in-text similarity, or semantic similarity. The score of similarity is usually being calculated using the distance between concepts and terms \autocite{diao2017homographic, kamal2019self, shahaf2015inside, sjobergh2007recognizing, indurthi2017fermi, mihalcea2010computational, yatsu2018comparison, zhang2017investigations, vadehra2017uwav}. 
\end{itemize}
\end{itemize}

\subsection{Algorithms}
\label{subsec:algorithms}
There are four types of algorithms in order to classify the text into humorous or not humorous:

\begin{itemize}
\item \textbf{\underline{Supervised Learning Approaches}} - Algorithms such as: Support Vector Machine (SVM), Random Forests (RF), Naive Bayes (NB), Decision Trees (DT), Logistic
Regression (LR), and K-Nearest Neighbor (K-NN) were the building blocks for the rudimentary classical NLP \autocite{mahajan2020humor, castro2016joke}, in the \autocite{jaiswal2019pun, hossain2019president} papers they had found RF was the best among these algorithms.  
\item \textbf{\underline{Deep Learning Approaches}} - Deep Learning (DL) is any type of neural network with more than 3 layers. The big advantage of this algorithm over the previous group of algorithms: it extracts features automatically without need of human intervention \autocite{shukla2019automatic}. DL contains various types of algorithms, such as: Convolutional Neural Network (CNN), Graph Convolutional Network (GCN), Recurrent Neural Network (RNN), Bidirectional RNN (BiRNN), Long Short-Term Memory Networks (LSTM), Bidirectional LSTM (BiLSTM), Gated Recurrent Unit (GRU), Bidirectional GRU (BiGRU), Contextual Memory Fusion Network (C-MFN), Multi-Layer Perceptron (MLP). DL Algorithms found to be very good in comprehensive tasks according to \autocite{alzubaidi2021review}. The Best algorithm according to \autocite{review2024until2022} that was found to be is LSTM. There were researchers that tried to combine two of algorithms above in order to get improvement \autocite{bertero2016multimodal, diao2020crga, diao2020homographic, fan2020phonetics, fan2020humor, ren2021abml}.
\item \textbf{\underline{Transfer Learning Approaches}} - The complexity of humor as a human emotional response necessitates extensive background knowledge and profound contextual awareness. As a result, transfer learning strategies that utilize pre-trained language models (PLMs) have become increasingly prominent in contemporary research, driven by their substantial progress in neural architecture development. This approach represents a machine learning framework that adapts existing PLMs for new applications, drawing upon their comprehensive knowledge base derived from common data across various fields \autocite{agarwal2020transfer}. the review of \autocite{review2024until2022} checked the following PLMs: Google’s BERT (Bidirectional
Encoder Representations from Transformers) \autocite{weller2019humor, cao2021self, morishita2020hitachi, akbar2021deep, mittal2021so, patro2021multimodal}, Google’s ALBERT (A Lite BERT) \autocite{song2021deepblueai},
Google’s XLNet (Generalized Autoregressive Pretraining for Language Understanding) \autocite{morishita2020hitachi},
Google’s Transformer-XL (Attentive Language Models Beyond a Fixed-Length Context) \autocite{morishita2020hitachi},
Facebook’s RoBERTa (Robustly Optimized BERT Pretraining Approach) \autocite{cao2021self, faraj2021sarcasmdet, mittal2021so, song2021deepblueai}, Facebook’s XLM (Enhancing BERT for Cross-lingual Language Model) \autocite{morishita2020hitachi, mittal2021so}, Microsoft’s CodeBERT, and OpenAI’s GPT-2 \autocite{morishita2020hitachi}. 
BERT was the most common model, but also the State-of-the-Art (SOTA) at the time. 
\autocite{akbar2021deep} showed that BERT was best in classifying humorous or humorless in text that was generated by GPT-2. Also, in SemEval-2020 the Hitachi team showed RoBERTa and BERT-large were the best models \autocite{faraj2021sarcasmdet}. 
\item \textbf{\underline{Rule-Based Approaches}} - Rule-based methodologies identify humorous elements within textual content through predefined criteria that utilize dictionaries and word databases. These criteria articulate the discovery of patterns and meaningful connections among data sample instances. The most common Algorithm being explored is the Lesk Algorithms, because it works pretty well and is pretty amicable towards more complex approaches based on it \autocite{miller2015automatic}. Additional computational models employed within the reviewed research include: Pointwise Mutual Information (PMI), dependency networks, Markov modeling systems, Word Sense Disambiguation (WSD) frameworks, Weighted Finite-State Transducers (WFSTs), Bayesian classification, Latent Semantic Analysis (LSA), Gloss Vector representations, and Inverse Document Frequency (IDF) techniques. However, this technique was compared to classic ML Algorithms and was found to be inferior \autocite{mikhalkova2017punfields, mihalcea2010computational}
\end{itemize}

\subsection{The Novelty of Our Study}
GPT-3 is very limited in generating jokes. According to \autocite{jentzsch2023chatgpt} it probably repeats the joke it memorized from human data (25 of his jokes are 90 percent of the generated jokes). It can explain a joke, but it is lacking since he mostly understands Puns/Wordplay but no other types of humor. Also, when it asks to explain a regular sentence (non-joke)
it classifies it as a joke and hallucinates on the explanation.

While GPT-4 is supposed to be better in generating jokes, it still struggles to generate a whole stand-up or comedy show according to \autocite{mirowski2024robot}, this is because the model is instructed to avoid certain words which can be a punchline. Also, it lacks the context of who the speaker is and who the audience is. Another issue we should bear in mind, that in order to create humor we need to make the opposite process of Chain-of-Thought (CoT). Instead of writing down each step at a time, such as in reasoning tasks, we need to skip steps and to leave to human a gap of information he or she should understand by themselves. This process called Leap-of-Thoguht (LoT), and it can't be done using prompting (unlike CoT). Creative Leap-of-Thought (CLoT) is an LLM LoRA tuned on a game of humor and association (Oogiri-GO dataset - in the language of English, Chinese, and Japanese), with text and images. and is found to outperform each model to the time of the publishment of this paper. \autocite{zhong2024let}

In the last three years, LLMs bloomed, especially Decoders, but also Encoders. In this paper we will check which models are better using classification to six classes - five types of jokes, and one not-funny sentence. The same as in \autocite{kasu2025deceptive}, a paper that used generated jokes by GPT-4o and showed that these jokes can be classified pretty good using Encoders. We will show that today, it can be done also for human ("real") jokes. And by a Fine-Tuned Decoder which is equal to making the comparisons to the best Encoder so far (RoBERTa). Nevertheless, certain Encoders such as RoBERTa-base and RoBERTa-large are a bit better than the Fine-Tuned Decoder (GPT-4o), but not significantly. 

\section{Methodology}
\subsection{Data Collection}

We collected English humor data categorized into five types, each defined below along with the number of examples and their respective sources.

\begin{figure}[H]
\centering
\resizebox{0.9\textwidth}{!}{%
\begin{tikzpicture}[
    node distance=1.2cm,
    box/.style={rectangle, draw=black, very thick, rounded corners=6pt, fill=blue!20, minimum width=3cm, minimum height=1cm, text centered, font=\scriptsize\bfseries},
    orangebox/.style={rectangle, draw=black, very thick, rounded corners=6pt, fill=orange!30, minimum width=3cm, minimum height=1cm, text centered, font=\scriptsize\bfseries},
    arrow/.style={thick,->,>=stealth}
]

\node[box, minimum width=4.5cm, minimum height=1.2cm] (sources) at (0,0) {
    \begin{tabular}{c}
    \textbf{Humor Sources} \\
    \textbf{(Web, Reddit)}
    \end{tabular}
};

\node[box, minimum width=6cm, minimum height=2cm, below=of sources] (collection) {
    \begin{tabular}{c}
    \textbf{Data Collection} \\
    \textbf{by Categories} \\[0.2cm]
    \scriptsize Absurdity \quad Dark \quad Irony \\
    \scriptsize Social Commentary \quad Wordplay
    \end{tabular}
};

\node[box, minimum width=7cm, minimum height=2.2cm, below=of collection] (filtering) {
    \begin{tabular}{c}
    \textbf{Manual Filtering} \\[0.1cm]
    \scriptsize Jokes in categories 'Absurdity', 'Dark', 'Irony', \\
    \scriptsize 'Social Commentary' were Manual Filtering \\
    \scriptsize from jokes that also contain "Wordplay"
    \end{tabular}
};

\node[box, minimum width=2.8cm, minimum height=1.8cm, below left=1.5cm and 1.5cm of filtering] (humor) {
    \begin{tabular}{c}
    \textbf{Final Humor} \\
    \textbf{Dataset} \\
    \textbf{(5 Categories)}
    \end{tabular}
};

\node[orangebox, minimum width=2.8cm, minimum height=2.5cm, below right=1.5cm and 1.5cm of filtering] (nojoke) {
    \begin{tabular}{c}
    \textbf{No-Joke Class} \\
    \textbf{from Kaggle} \\
    \scriptsize (Negative Examples) \\[0.1cm]
    \scriptsize (200K Short Texts \\
    \scriptsize for Humor Detection) \\
    \scriptsize Same size as number of jokes
    \end{tabular}
};

\node[box, minimum width=4cm, minimum height=1.5cm, below=2.8cm of filtering] (final) {
    \begin{tabular}{c}
    \textbf{Final Dataset} \\
    \textbf{(6 Categories)}
    \end{tabular}
};

\draw[arrow] (sources) -- (collection);
\draw[arrow] (collection) -- (filtering);
\draw[arrow] (filtering) -- (humor);
\draw[arrow] (filtering) -- (nojoke);
\draw[arrow] (humor) -- (final);
\draw[arrow] (nojoke) -- (final);

\draw[arrow] (humor.east) -- (nojoke.west);
\draw[arrow] (nojoke.west) -- (humor.east);

\end{tikzpicture}
}%
\caption{Data Collection and Processing Pipeline}
\label{fig:data-pipeline}
\end{figure}
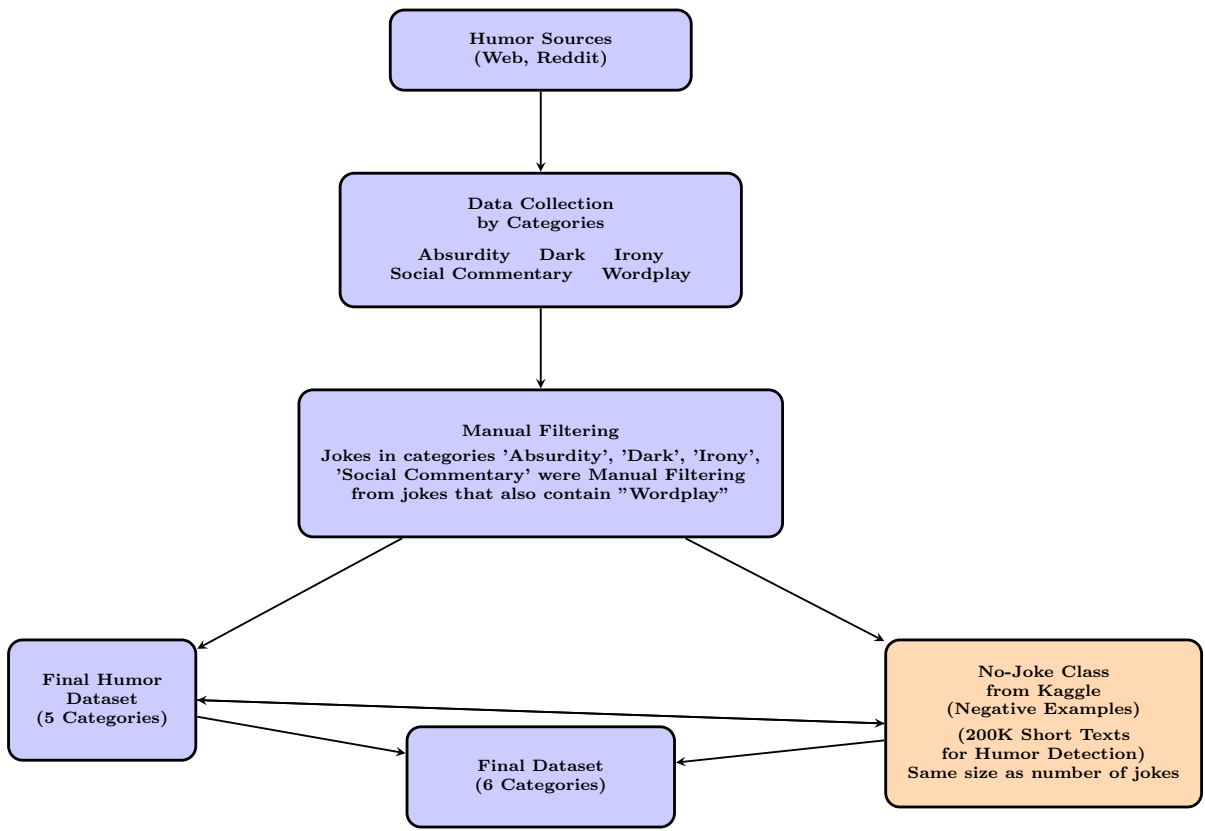

\textbf{Absurdity Jokes}—humor based on nonsensical or illogical scenarios that defy common sense—were sourced from \autocite{upjoke_absurd} (58 jokes), \autocite{rd_absurd} (13), \autocite{reddit_absurd} (21), and \autocite{today_absurd} (24).

\textbf{Dark Jokes}—humor involving taboo, morbid, or tragic subjects presented in a humorous way—were scraped from \autocite{coolist_dark} (115).

\textbf{Irony Jokes}—humor that relies on stating the opposite of what's expected or highlighting contradictions between expectations and reality—were collected from \autocite{discover_irony} (75) and \autocite{boredpanda_irony} (73).

\textbf{Wordplay Jokes}—humor derived from clever manipulation, puns, or play on words—were taken from \autocite{inews_wordplay} (105) and \autocite{styler_wordplay} (273).

\textbf{Social Commentary Jokes}—humor highlighting societal issues, trends, or behaviors in a satirical or critical manner—were gathered from \autocite{upjoke_commentary} (11), \autocite{laughfactory_political} (24), \autocite{rd_political} (29), and \autocite{rd_presidential} (2).

\subsection{Data Preprocessing}
\subsubsection{Clean Humor Ambiguity Types}
All jokes have been checked one after another manually, and any joke from a non-wordplay category that also contained a wordplay element, was excluded from the dataset. This step was taken to ensure the classification task remains a multi-class problem rather than a multi-label one. However, to illustrate the inherent ambiguity in humor categorization, we present in Table~\ref{tab:mixed-with-wordplay-humor-examples} several representative jokes that feature multiple humor types, especially where wordplay overlaps with others.

\begin{table}[H]
\centering
\begin{tabular}{|p{0.3\linewidth}|p{0.2\linewidth}|p{0.45\linewidth}|}
\hline
\textbf{Joke} & \textbf{Humor Types} & \textbf{Explanation} \\
\hline
\small
A high school senior visits a psychic... "I've applied to 10 different colleges," the student said. "Which ones will accept me? Which one will I attend?" "That is hard to say," said the psychic. "But you will spend an absurd sum of money." "How do you know this?" the student asked. The psychic replied, "It's mostly intuition." 
& \small Absurdity, Wordplay 
& \small Absurd: The student consults a psychic for college decisions, an illogical scenario. Wordplay: The punchline hinges on the pun between "intuition" (gut feeling) and "in tuition" (financial cost). \\
\hline
\small
It turns out a major new study recently found that humans eat more bananas than monkeys. It's true. I can't remember the last time I ate a monkey.
& \small Dark, Wordplay 
& \small Wordplay: The joke plays on the ambiguous sentence structure, twisting the meaning from a comparison of diet to a claim about eating monkeys. Dark: It introduces a taboo subject—eating primates—in a casual and humorous tone. \\
\hline
\small
It is funny and sad thing how a group of squid is not called a squad.
& \small Irony, Wordplay 
& \small Wordplay: Plays on the phonetic similarity between "squid" and "squad," setting up a pun. Irony: Highlights the mismatch between what would be a fitting name and actual naming conventions, provoking humor through unexpected linguistic rules. \\
\hline
\small
If con is the opposite of pro, then is Congress the opposite of progress?
& \small Social Commentary, Wordplay
& \small Wordplay: The joke plays on the prefixes "con" and "pro," twisting the meaning of "Congress" into an opposite of "progress." Social Commentary: Critiques political inefficiency or dysfunction in a satirical way. \\
\hline
\end{tabular}
\caption{Examples of different humor types mixed with wordplay humor - that were removed from our dataset}
\label{tab:mixed-with-wordplay-humor-examples}
\end{table}

\subsubsection{Replace repeating words}
Repetition of the same word or a collocation in the same category were replaced by random words. For example, the collocation "the ironic person" repeated itself in the data of the irony jokes, therefore I replaced it with a random first and last name. 
e.g.: 

\colorbox{red!20}{\parbox{\textwidth}{\textbf{Before:} ``Why did \textit{the ironic person} wear sunscreen? Because they wanted to get a sunburn!''}}

\colorbox{green!20}{\parbox{\textwidth}{\textbf{After:} ``Why did \textit{Kelly Jones} wear sunscreen? Because \textit{she} wanted to get a sunburn!''}}

\subsubsection{Replace the category type words with other words}
Words that suggest a clue to the category were changed. For example, half of the jokes in the irony category included the word: "ironic" or "irony", we changed these words with funny/surprising/ or any other suiting word for the context. e.g.: 

\colorbox{red!20}{\parbox{\textwidth}{\textbf{Before:} "What's \textit{ironic} about the Bible? In one of the most interesting \textit{irony} examples, the most shoplifted book in America is The Bible."}}

\colorbox{green!20}{\parbox{\textwidth}{\textbf{After:} "What's \textit{funny} about the Bible? In one of the most interesting \textit{absurd} examples, the most shoplifted book in America is The Bible."}}

\subsection{Negative Examples}
Regular sentences (not funny / non-joke) were randomly sampled from a Kaggle dataset \autocite{kaggle_humor}, and we made sure that the amount of regular sentences is equal to the amount of all other jokes.

\begin{table}[H]
    \centering
    \begin{tabular}{|c|c|}
        \hline
        Type of Sentence & Number of Examples \\
        \hline
        Absurdity  & 75 \\
        Dark & 77 \\
        Irony & 105 \\
        Wordplay & 378 \\
        Social Commentary & 62 \\
        No-Joke (regular sentence) & 697 \\
        Total & 1392 \\
        \hline
    \end{tabular}
    \caption{Number of examples for each Category after cleaning}
    \label{tab:sample}
\end{table}

\subsection{Train, Test, Validation Data Split}
The data were split into: Train (80\%), Validation (10\%), Test (10\%). Stratify was used to maintain the same ratio between labels within category groups. 

\subsection{Training}
\begin{figure}[H]
\centering
\begin{tikzpicture}
\label{tab:EncodersFineTuning}
\begin{axis}[
  width=14cm,
  height=9.5cm,
  xlabel={Training Step},
  ylabel={Loss},
  title={Encoders Training Loss Over Time},
  legend style={at={(1,1)}},
  grid=major,
  xmin=0,
  ymin=0,
  ymax=1.4,
]
\addplot[smooth, thick, color=blue] coordinates {
(0, 1.0029) (2, 0.7404) (4, 0.6004) (6, 0.5267) (8, 0.4623)
(10, 0.3804) (12, 0.2637) (14, 0.2065) (16, 0.1626) (18, 0.1314)
(20, 0.1199) (22, 0.0892) (24, 0.0694) (26, 0.0396) (28, 0.0336)
(30, 0.0158) (32, 0.0108) (34, 0.0081) (36, 0.0073) (38, 0.0071)
};
\addlegendentry{BERT-base-uncased}

\addplot[smooth, thick, color=red] coordinates {
(0, 1.0902) (2, 0.5769) (4, 0.4166) (6, 0.2756) (8, 0.1784)
(10, 0.1038) (12, 0.0535) (14, 0.0312) (16, 0.0224) (18, 0.015)
(20, 0.0118) (22, 0.0093) (24, 0.0078) (26, 0.0073) (28, 0.0061)
(30, 0.005) (32, 0.0044) (34, 0.0043) (36, 0.0038) (38, 0.0034)
};
\addlegendentry{DistilBERT-base-uncased}

\addplot[smooth, thick, color=green!60!black] coordinates {
(0, 0.8509) (2, 0.5178) (4, 0.331) (6, 0.196) (8, 0.1177)
(10, 0.1023) (12, 0.0429) (14, 0.0228) (16, 0.0058) (18, 0.0034)
(20, 0.0029) (22, 0.0023) (24, 0.002) (26, 0.0018) (28, 0.0016)
(30, 0.0015) (32, 0.0015) (34, 0.0014) (36, 0.0014) (38, 0.0014)
};
\addlegendentry{mBERT-base-cased}

\addplot[smooth, thick, color=orange] coordinates {
(0, 0.8006) (2, 0.4713) (4, 0.2934) (6, 0.1849) (8, 0.1242)
(10, 0.0845) (12, 0.0392) (14, 0.0199) (16, 0.0099) (18, 0.0046)
(20, 0.0021) (22, 0.0014) (24, 0.0011) (26, 0.0008) (28, 0.0007)
(30, 0.0006) (32, 0.0006) (34, 0.0005) (36, 0.0005) (38, 0.0005)
};
\addlegendentry{ALBERT-base-v2}

\addplot[smooth, thick, color=purple] coordinates {
(0, 1.0947) (2, 0.622) (4, 0.4725) (6, 0.3249) (8, 0.222)
(10, 0.1481) (12, 0.0965) (14, 0.0598) (16, 0.0304) (18, 0.013)
(20, 0.0131) (22, 0.0067) (24, 0.0067) (26, 0.0041) (28, 0.0022)
(30, 0.0014) (32, 0.0013) (34, 0.0013) (36, 0.0012) (38, 0.0012)
};
\addlegendentry{DeBERTa-v3-base}

\addplot[smooth, thick, color=teal] coordinates {
(0, 0.9886) (2, 0.6086) (4, 0.3795) (6, 0.2657) (8, 0.182)
(10, 0.1089) (12, 0.0855) (14, 0.0297) (16, 0.0191) (18, 0.0089)
(20, 0.0045) (22, 0.0015) (24, 0.0014) (26, 0.0012) (28, 0.0012)
(30, 0.001) (32, 0.0009) (34, 0.0009) (36, 0.0009) (38, 0.0008)
};
\addlegendentry{RoBERTa-base}
\addplot[smooth, thick, color=magenta] coordinates {
(0, 1.1001) (2, 0.6845) (4, 0.4445) (6, 0.2793) (8, 0.1463)
(10, 0.0683) (12, 0.0524) (14, 0.026) (16, 0.0071) (18, 0.0013)
(20, 0.0002) (22, 0.0002) (24, 0.0002) (26, 0.0002) (28, 0.0002)
(30, 0.0002) (32, 0.0001) (34, 0.0001) (36, 0.0001) (38, 0.0001)
};
\addlegendentry{RoBERTa-large}

\addplot[smooth, thick, color=brown] coordinates {
(0, 1.2027) (2, 0.7245) (4, 0.5943) (6, 0.4405) (8, 0.306)
(10, 0.2195) (12, 0.1553) (14, 0.112) (16, 0.0634) (18, 0.0349)
(20, 0.021) (22, 0.0166) (24, 0.007) (26, 0.0028) (28, 0.0019)
(30, 0.0017) (32, 0.0017) (34, 0.0015) (36, 0.0061) (38, 0.0013)
};
\addlegendentry{XLM-RoBERTa-base}

\addplot[smooth, thick, color=cyan] coordinates {
(0, 0.8758) (2, 0.4551) (4, 0.2942) (6, 0.1733) (8, 0.1073)
(10, 0.0712) (12, 0.0304) (14, 0.0236) (16, 0.009) (18, 0.0013)
(20, 0.001) (22, 0.001) (24, 0.005) (26, 0.0004) (28, 0.0003)
(30, 0.0005) (32, 0.0004) (34, 0.0003) (36, 0.0003) (38, 0.0002)
};
\addlegendentry{XLNet-base-cased}

\addplot[smooth, thick, color=lime] coordinates {
(0, 0.7825) (2, 0.3632) (4, 0.1991) (6, 0.0415) (8, 0.0056)
(10, 0.0004) (12, 0.0001) (14, 0.0135) (16, 0.0042) (18, 0.0018)
(20, 0.0013) (22, 0.0008) (24, 0.0006) (26, 0.0005) (28, 0.0005)
(30, 0.0004) (32, 0.0004) (34, 0.0004) (36, 0.0004) (38, 0.0003)
};
\addlegendentry{ModernBERT-base}

\addplot[smooth, thick, color=violet] coordinates {
(0, 1.0018) (2, 0.5543) (4, 0.3462) (6, 0.1954) (8, 0.1092)
(10, 0.0567) (12, 0.0307) (14, 0.0581) (16, 0.0431) (18, 0.0223)
(20, 0.0128) (22, 0.0053) (24, 0.004) (26, 0.0027) (28, 0.002)
(30, 0.0013) (32, 0.0011) (34, 0.0007) (36, 0.0006) (38, 0.0006)
};
\addlegendentry{bert-large-uncased}

\addplot[smooth, thick, color=gray] coordinates {
(0, 1.0018) (2, 0.6221) (4, 0.4167) (6, 0.2838) (8, 0.1871)
(10, 0.1259) (12, 0.0879) (14, 0.0581) (16, 0.1169) (18, 0.0938)
(20, 0.0457) (22, 0.0372) (24, 0.0269) (26, 0.006) (28, 0.0012)
(30, 0.0027) (32, 0.0009) (34, 0.0003) (36, 0.0003) (38, 0.0003)
};
\addlegendentry{albert-large-v2}

\addplot[smooth, thick, color=olive] coordinates {
(0, 1.2084) (2, 0.8536) (4, 0.7139) (6, 0.5697) (8, 0.4303)
(10, 0.3191) (12, 0.2389) (14, 0.1781) (16, 0.1169) (18, 0.0938)
(20, 0.0457) (22, 0.0372) (24, 0.0269) (26, 0.006) (28, 0.0012)
(30, 0.0027) (32, 0.0009) (34, 0.0003) (36, 0.0003) (38, 0.0003)
};
\addlegendentry{DeBERTa-v3-large$\circ\ast\ast$}
\addplot[smooth, thick, color=darkgray] coordinates {
(0, 0.9809) (2, 0.6389) (4, 0.7139) (6, 0.5697) (8, 0.4303)
(10, 0.3191) (12, 0.2389) (14, 0.1781) (16, 0.1169) (18, 0.0938)
(20, 0.0457) (22, 0.0372) (24, 0.0269) (26, 0.006) (28, 0.0012)
(30, 0.0027) (32, 0.0009) (34, 0.0003) (36, 0.0003) (38, 0.0003)
};
\addlegendentry{XLM-RoBERTa-large}
\addplot[smooth, thick, color=black] coordinates {
(0, 0.8139) (2, 0.4905) (4, 0.2427) (6, 0.1051) (8, 0.0318)
(10, 0.0055) (12, 0) (14, 0.0003) (16, 0.0002) (18, 0.0002)
(20, 0.0002) (22, 0.0001) (24, 0.0001) 
};
\addlegendentry{NeoBERT$\circ\ast$}
\addplot[smooth, thick, color=red!80!black] coordinates {
(0, 0.6033)
(2, 0.1772)
(4, 0.0485)
(6, 0.0095)
(8, 0.0063)
(10, 0)
};
\addlegendentry{ModernBERT-large}
\addplot[smooth, thick, color=green!80!black] coordinates {
(0, 0.9809)
(2, 0.6389)
(4, 0.7139)
(6, 0.5697)
(8, 0.4303)
(10, 0.3191)
(12, 0.2389)
(14, 0.1781)
(16, 0.1169)
(18, 0.0938)
(20, 0.0457)
(22, 0.0372)
(24, 0.0269)
(26, 0.006)
(28, 0.0012)
(30, 0.0027)
(32, 0.0009)
(34, 0.0003)
(36, 0.0003)
(38, 0.0003)
};
\addlegendentry{XLNet-large-cased$\ast$}
\end{axis}
\end{tikzpicture}
\caption{Training loss of models over time (steps)}
\end{figure}
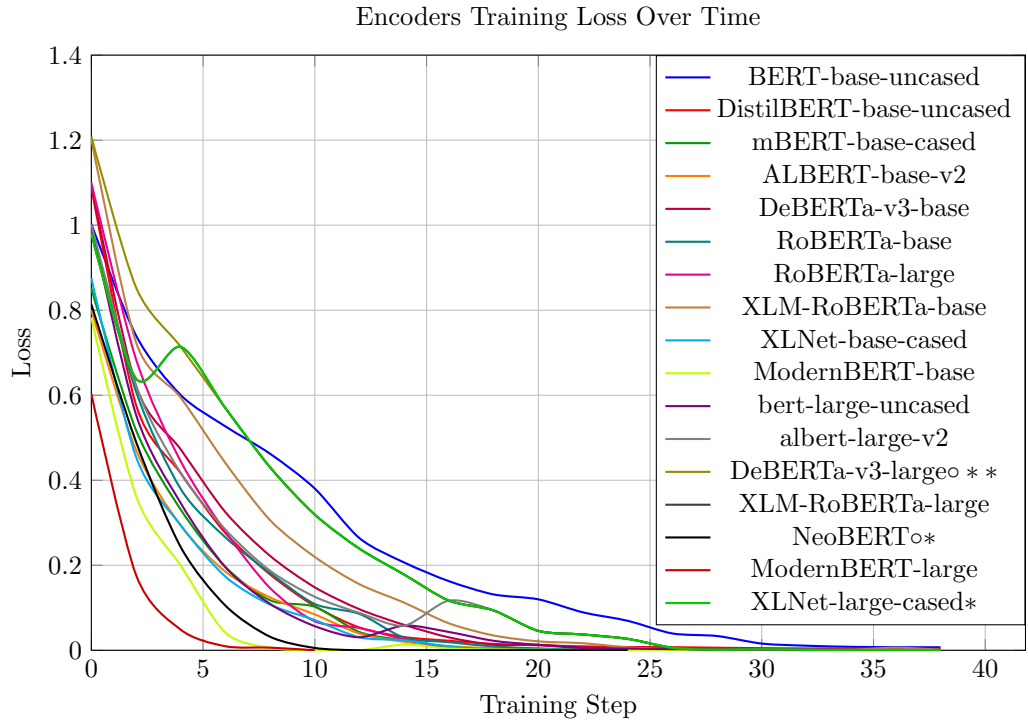

\begin{flushleft}
\footnotesize{$^{\ast}$Trained with batch size 8 instead of 16.
}\\
\footnotesize{$^{\ast\ast}$Trained with batch size 4 instead of 16.
}\\
\footnotesize{$^{\circ}$Model Quantization was set to be bf16 precision instead of fp32
}
\end{flushleft}
\vspace{0.5em}

\paragraph{\textnormal{Fine-Tuning was done to all Encoder Models, an example for a fine-tuning with seed 42 is available at }\hyperref[tab:EncodersFineTuning]{Figure~2}, \textnormal{using HuggingFace and PyTorch (transformers). The encoders were fine-tuned $3$ times for $20$ epochs, batch size was 16 (except XLNet-large-cased and NeoBERT which had batch size of 8, DeBERTa-v3-large with batch size of 4), all models were full precision fp32 (except DeBERTa-v3-large and NeoBERT which used bf16). The runs were done with 3 constant seeds ($42$, $1337$, $2025$), so the models are reproducible. In each time when training loss was close enough to zero (epsilon of 0.0001) then training was stopped. In each time the selected encoder was chosen according to its epoch with maximum f1-macro score (this metric was chosen since the data categories are biased). Eventually all 3 models from each run, were averaged, paving the way for a non-normal distribution with mean and standard-deviation. Also we had average for each run the recorded (best epoch according to f1-macro) metrics such as: Recall-Macro, Precision-Macro, and Accuracy.}}

\paragraph{\textnormal{While BART (Encoder2Decoder model) was checked only with zero shot learning (few-shots learning isn't available for this model). Flan-T5 was checked both for zero-shot learning and few-shots.}}

\begin{figure}[H]
\begin{tikzpicture}
\label{tab:FigureGPTFineTuning}
\begin{axis}[
  width=\textwidth,
  height=5cm,
  xlabel={Training Step},
  ylabel={Loss Value},
  title={Training and Validation Loss for GPT-4o Fine-Tuning},
  grid=both,
  grid style={line width=.1pt, draw=gray!10},
  major grid style={line width=.2pt,draw=gray!50},
  legend style={at={(0.98,0.85)}, anchor=north east},
  ymin=0,
  ymax=8,
  xmin=0,
  xmax=850,
  mark size=0.4pt,
]

\addplot[color=blue, mark=*, mark size=0.3pt, line width=0.8pt, opacity=0.7] coordinates {
    (1, 6.674) (56, 0.859) (169, 0.538) (324, 0.965) (429, 0.62) (812, 0.979)
};

\addplot[color=red, mark=square*, mark size=0.4pt, line width=1pt] coordinates {
    (10, 3.949) (30, 0.003) (70, 0.013) (140, 0.383) (300, 0.046) (590, 0.947) (730, 0.001)
};

\legend{Training Loss, Validation Loss}
\end{axis}
\end{tikzpicture}
\caption{Training and validation loss progression for GPT-4o fine-tuning}
\end{figure}
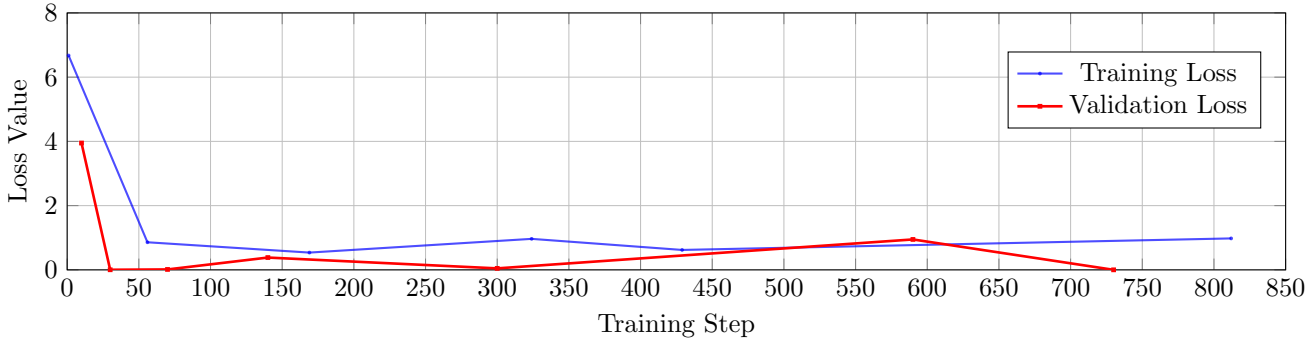

GPT-4o (the main/big decoder) was also Fine-Tuned, using OPENAI API, it was trained for 3 epochs, batch size of 4 examples each time, and a Learning Rate (LR) Multiplier of 2. we had found that when GPT-4o decoder was fine-tuned it got very fast to zero training loss, which means the model is over-fitting and "hungry" for data \hyperref[tab:FigureGPTFineTuning]{Figure~3}.

\section{Results}
\begin{table}[H]
    \centering
    \label{tab:TableResults}
    \resizebox{\textwidth}{!}{
    \begin{tabular}{|l|c|c|c|c|}
    \hline
    \textbf{Model Type \& Name} & \textbf{Recall-Macro} & \textbf{Precision-Macro} & \textbf{F1-Macro} & \textbf{Accuracy} \\
    \hline
    \multicolumn{5}{|l|}{\textbf{Encoders}} \\
    \texttt{BERT-base-uncased} & $0.5635 \pm 0.0601$ & $0.6135 \pm 0.0606$ & $0.5808 \pm 0.0588$ & $0.8024 \pm 0.0251$ \\
    \texttt{BERT-base-multilingual} & $0.7994 \pm 0.045$ & $0.7492 \pm 0.0476$ & $0.7674 \pm 0.0481$ & $0.881 \pm 0.027$ \\
    \texttt{XLNet-base-cased} & $0.7455 \pm 0.0471$ & $0.7185 \pm 0.0362$ & $0.7212 \pm 0.0402$ & $0.8642 \pm 0.0214$ \\
    \texttt{distilBERT-base-uncased} & $0.7682 \pm 0.0159$ & $0.7274 \pm 0.03679$ & $0.7358 \pm 0.0333$ & $0.8619 \pm 0.027$ \\
    \texttt{ALBERT-base} & $0.7579 \pm 0.0789$ & $0.7402 \pm 0.0409$ & $0.7201 \pm 0.053$ & $0.8476 \pm 0.023$ \\
    \texttt{RoBERTa-base} & $\boldsymbol{0.8745 \pm 0.0081}^{\dag}$ & $\boldsymbol{0.8652 \pm 0.0304}^{\dag}$ & $\boldsymbol{0.8566 \pm 0.0164}^{\dag}$ & $\boldsymbol{0.9286 \pm 0.0124}^{\dag}$ \\
    \texttt{XLM-roBERTa-base} & $0.8061 \pm 0.042$ & $0.7735 \pm 0.057$ & $0.7462 \pm 0.0569$ & $0.8691 \pm 0.0393$ \\
    \texttt{DeBERTa-v3-base} & $0.7855 \pm 0.0691$ & $0.784 \pm 0.1056$ & $0.7733 \pm 0.0882$ & $0.8905 \pm 0.0406$ \\
    \texttt{ModernBERT-base} & $0.687 \pm 0.0131$ & $0.722 \pm 0.0484$ & $0.6871 \pm 0.0182$ & $0.8642 \pm 0.0124$ \\
    \texttt{BERT-large-uncased} & $0.7187 \pm 0.0549$ & $0.688 \pm 0.0393$ & $0.6959 \pm 0.0425$ & $0.8381 \pm 0.0165$ \\
    \texttt{XLNet-large-cased}$^{\ast}$ & $0.7936 \pm 0.0628$ & $0.7451 \pm 0.048$ & $0.7606 \pm 0.0541$ & $0.8714 \pm 0.0247$ \\
    \texttt{ALBERT-large-v2} & $0.7119 \pm 0.0133$ & $0.7101 \pm 0.0177$ & $0.7057 \pm 0.0129$ & $0.8524 \pm 0.0109$ \\
    \texttt{RoBERTa-large} & $0.8594 \pm 0.0473^{\dag}$ & \underline{$0.8501 \pm 0.0318$}$^{\dag}$ & $0.8503 \pm 0.0355^{\dag}$ & $\boldsymbol{0.9286 \pm 0.0071}^{\dag}$ \\
    \texttt{XLM-roBERTa-large} & $0.8514 \pm 0.0348$ & $0.8228 \pm 0.0519$ & $0.8308 \pm 0.0441$ & $0.9214 \pm 0.0189$ \\
    \texttt{DeBERTa-v3-large}$^{\circ}$$^{\ast\ast}$ & $0.8441 \pm 0.0271$ & $0.8034 \pm 0.0378$ & $0.8145 \pm 0.0304$ & $0.9048 \pm 0.0.0109$ \\
    \texttt{ModernBERT-large} & $0.7806 \pm 0.0195$ & $0.7864 \pm 0.0539$ & $0.7611 \pm 0.0151$ & $0.8929 \pm 0.0071$ \\
    \texttt{NeoBERT}$^{\circ}$$^{\ast}$ & $0.7719 \pm 0.0861$ & $0.8191 \pm 0.0274$ & $0.7765 \pm 0.0786$ & $0.9049 \pm 0.0251$ \\
    \hline
    \multicolumn{5}{|l|}{\textbf{Encoder-Decoder}} \\
    \texttt{BART-large-mnli-zero shot} & $0.2488$ & $0.2541$ & $0.1841$ & $0.2071$ \\
    \texttt{Flan-T5-base-zero shot} & $0.2292$ & $0.2843$ & $0.1575$ & $0.3696$ \\
    \texttt{Flan-T5-base-few shots} & $0.1674$ & $0.1619$ & $0.1079$ & $0.3254$ \\
    \hline
    \multicolumn{5}{|l|}{\textbf{Decoders}} \\
    \texttt{Llama-3.2-3B-Instruct-zero shot} & $0.3600$ & $0.2934$ & $0.1729$ & $0.1929$ \\
    \texttt{Gemma-2-2b-it-zero shot} & $0.3697$ & $0.4944$ & $0.3665$ & $0.6571$ \\
    \texttt{Qwen2-7B-Instruct-zero shot} & $0.3746$ & $0.4221$ & $0.2759$ & $0.2595$ \\
    \texttt{Mistral-7B-Instruct-v0.2-zero shot} & $0.4055$ & $0.5071$ & $0.3184$ & $0.3000$ \\
    \texttt{GPT-4-zero shot} & $0.5919$ & $0.5808$ & $0.5047$ & $0.5214$ \\
    \texttt{Llama-3.2-3B-Instruct-few shots} & $0.3895$ & $0.4328$ & $0.2086$ & $0.2143$ \\
    \texttt{Gemma-2-2b-it-few shots} & $0.4234$ & $0.4945$ & $0.3759$ & $0.4786$ \\
    \texttt{Qwen2-7B-Instruct-few shots} & $0.4791$ & $0.4777$ & $0.3452$ & $0.2950$ \\
    \texttt{Mistral-7B-Instruct-v0.2-few shots} & $0.4801$ & $0.6030$ & $0.3865$ & $0.3857$ \\
    \texttt{GPT-4-few shots} & $0.6381$ & $0.6878$ & $0.5955$ & $0.6429$ \\
    \texttt{GPT-4o-fine-tuned} & \underline{$0.8618 \pm 0.0025^{\dag}$} & $0.8464 \pm 0.0089^{\dag}$ & \underline{$0.8522 \pm 0.0056^{\dag}$} & \underline{$0.9238 \pm 0.0041^{\dag}$} \\
    \hline
    \end{tabular}
    }
    \caption{Recall-Macro, Precision-Macro, F1-Macro, and Accuracy test scores Among Many Different Models}
    \small{\textbf{Bold} scores represent the best score in column. \\
    \underline{Underlined} scores are the second-best score in the column.}
\end{table}

\begin{flushleft}
\footnotesize{$\pm$ add or subtract standard deviation from mean}\\
\footnotesize{$^{\ast}$Trained with batch size 8 instead of 16.
}\\
\footnotesize{$^{\ast\ast}$Trained with batch size 4 instead of 16.
}\\
\footnotesize{$^{\circ}$Model Quantization was set to be bf16 precision instead of fp32
}\\
\footnotesize{$^{\dag}$ - no significant difference (p.value greater than 0.05) according to Welch’s t-test}
\end{flushleft}
\vspace{0.5em}

\paragraph{\textnormal{\hyperref[tab:TableResults]{Table~3} shows that larger encoders perform better than their "base" version (except ALBERT and RoBERTa), with a mean gap of 0.05 f1-macro score points (ALBERT and RoBERTa were also included in the average). However, it's unclear why RoBERTa and ALBERT had the same results between the different model sizes. We have also found RoBERTa-base to be the best model.}}

\paragraph{\textnormal{In the Encoders-to-Decoders and Decoders, we have checked zero-shot and few-shots learning. Flan-T5 had a surprising outcome: few-shots learning performed worse than zero-shot learning, this is probably because of the fact that digesting long prompt might be a problem for small models. All the models which were checked for zero-shot and few-shots learning had lagged behind the RoBERTa Encoder result.}}

\paragraph{\textnormal{However, in the crux of this study, we had run the fine-tuned GPT-4o 3 times in a greedy mode (temperature = 0), with a different seed for each run (42, 1337, 2025) and we got f1-macro score of 0.8522 which is the second best result and it stands in par with the best Encoder, RoBERTa (the Welch’s t-test showed there isn't a statistical difference between them).}}

\section{Conclusions}
This study aimed to probe whether various LLMs can understand Humor. The test was done by classifying each sentence to one of six categories (five types of humor or no-joke regular/negative sentence). The surprising result is that both fine-tuned Encoders and fine-tuned Decoders perform equally well. This contradicts the previous approach where Encoder (RoBERTa) was considered to be better.

\section{Limitations}
It's important to mention that issues such as heterogeneous data (which may imply different jargon for different categories), also the data was scarce with only 1394 examples (when half of them 697 are regular sentences, "Non-Joke"), the length of sentences in different categories, as well as semantic meaning that can help classification (e.g. Absurdity was mainly taken from the world of politics jokes) were not taken into consideration under this paper. Also, GPT-5 was only introduced in the recent couple of days, and GPT-5 is not publicly available as a distinct model for fine-tuning in the time of publication of this paper. 

\printbibliography{}

\end{document}